\documentclass[letterpaper, 10 pt, conference]{ieeeconf}  

\IEEEoverridecommandlockouts                              

\usepackage{graphicx} 
\usepackage{amsmath} 
\usepackage{amssymb}  
\usepackage[ruled,vlined,noend,linesnumbered]{algorithm2e}
\usepackage{comment}
\usepackage{subcaption}
\usepackage[font=small]{caption} 
\usepackage{svg}    
\usepackage[switch]{lineno}
\usepackage{bbm}

\newtheorem{lemma}{Lemma}
\newtheorem{theorem}{Theorem}
\DeclareMathOperator\erf{erf}
\newcommand*\diff{\mathop{}\!\mathrm{d}}

\title{\LARGE \bf
Planning on a (Risk) Budget: \\ Safe Non-Conservative Planning in Probabilistic Dynamic Environments
}

\author{Hung-Jui Huang$^{1}$, Kai-Chi Huang$^{1}$, Michal Čáp$^{1}$, Yibiao Zhao$^{1}$, Ying Nian Wu$^{2}$, Chris L. Baker$^{1}$
\thanks{$^{1}$ISEE AI, Cambridge, Massachusetts, USA}%
\thanks{$^{2}$University of California, Los Angeles, California, USA}
}

\begin{document}

\maketitle
\thispagestyle{empty}
\pagestyle{empty}

\begin{abstract}
Planning in environments with other agents whose future actions are uncertain often requires compromise between safety and performance. Here our goal is to design efficient planning algorithms with guaranteed bounds on the probability of safety violation, which nonetheless achieve non-conservative performance. To quantify a system's risk, we define a natural criterion called interval risk bounds (IRBs), which provide a parametric upper bound on the probability of safety violation over a given time interval or task. We present a novel receding horizon algorithm, 
and prove that it can satisfy a desired IRB.
Our algorithm maintains a dynamic risk budget which constrains the allowable risk at each iteration, and guarantees recursive feasibility by requiring a safe set to be reachable by a contingency plan within the budget.
We empirically demonstrate that our algorithm is both safer and less conservative than strong baselines in two simulated autonomous driving experiments in scenarios involving collision avoidance with other vehicles, and additionally demonstrate our algorithm running on an autonomous class 8 truck.

\end{abstract}

\section{Introduction}

Planning safe, efficient robot actions in the presence of other agents whose future behavior is uncertain is a challenging, unsolved problem.
Even in structured environments, where prior knowledge enables probabilistic prediction of agents' motions (e.g.~\cite{tang2019,cui2019,jacobs2019}), planning optimal feedback policies, subject to constraints bounding the probability of safety violation, is computationally intractable~\cite{khonji2019}. 
In practice, the optimal solution must be approximated, introducing a tradeoff between safety and performance: systems which satisfy a very high safety threshold may behave too conservatively, while systems which are non-conservative enough to meet practical efficiency requirements may not provide the desired safety guarantees, requiring close human supervision.
For challenging, safety-critical applications like autonomous driving, new techniques are needed for efficiently computing plans which guarantee hard constraints on safety, while achieving non-conservative performance.

Safety requirements are often quantified in terms of the maximum allowed probability of safety violation over a chosen time interval or task. In this paper, our goal is to enforce bounds of this form on planning algorithms, while still achieving non-conservative performance. We define a natural criterion for probabilistically safe planning, called interval risk bounds (IRBs), which provide a parametric upper bound on the probability of safety violation, or risk, during a given episode or time interval as a function of the interval length. We show that although computationally efficient open-loop approximations to the optimal feedback policy can satisfy an IRB over a finite interval, in practice their performance is too conservative for many applications. To improve performance, a feedback mechanism can be introduced using a receding horizon control (RHC) strategy. However, we show that --- surprisingly --- RHC algorithms which use joint chance constraints~\cite{ono2012} to reduce conservatism by non-uniformly allocating risk over the planning horizon can allocate the maximum allowed risk at every step of the policy in the worst case, making it difficult to provide meaningful bounds on the risk the policy will incur.

\begin{figure}[t]
\centering
\includegraphics[width=0.475\textwidth]{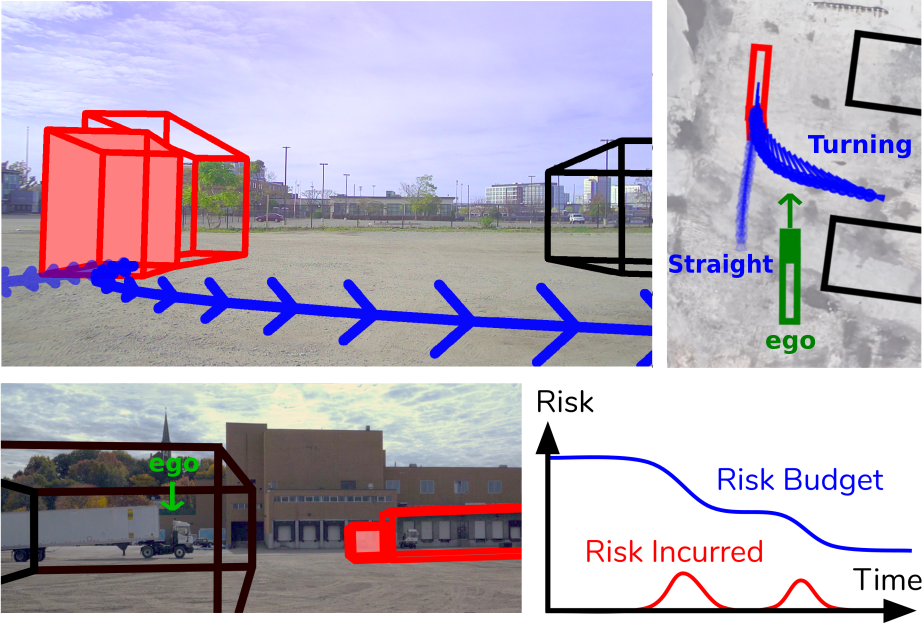}
\caption{{\em Top Left, Top Right, Bottom Left:} Our algorithm for safe planning with dynamic obstacles running on an autonomous tractor-trailer system, shown from first-person view, bird's eye view, and third person view; other truck (red), prediction (blue), static obstacles (black), ego truck (green). {\em Bottom Right:} We propose a novel algorithmic framework in which risk is provably bounded by a dynamic risk budget. Our algorithm achieves safe, non-conservative planning by constrained optimization of risk allocation over time.
}
\label{fig:page1}
\end{figure}


We present a novel RHC algorithm to address this issue. We prove theoretically that our algorithm can satisfy a given IRB by drawing a correspondence between the algorithmic parameters and those of the IRB, and we demonstrate empirically that our algorithm achieves safe and non-conservative performance in several autonomous driving experiments.
The algorithm maintains a risk budget which determines a joint chance constraint at each step of RHC, and which is updated after every step by subtracting the risk incurred. The joint chance constraint both allocates risk non-uniformly, reducing conservatism, and ensures recursive feasibility by requiring a safe set to be reachable by a contingency plan within the remaining risk budget. Although other definitions are possible, in this work we define our safe set in terms of \textit{passive safety}~\cite{macek2009}: the robot is assumed to be safe if it is not in collision with any obstacles, or if it is stopped.

Several factors enable our algorithm to achieve non-conservative performance. First, the contingency plan reduces conservatism by providing a closed-loop policy which handles all high-risk states, allowing a less conservative policy in the remaining low-risk states. Second, our algorithm uses probabilistic dynamics and agent models, which can reduce conservatism relative to worst-case assumptions by incorporating knowledge about the environment and agents' behavior patterns; we assume that these models have been learned  a priori and validated independently of our system. Lastly, our framework is compatible with planning heuristics for reducing conservativeness, allowing us to improve performance while ensuring safety.




We apply our framework in several experiments to support our theoretical arguments, and demonstrate the performance and practicality of our method. First, in a simplified interactive autonomous driving scenario, we show that our approach satisfies the desired risk bound, but does not undershoot due to over-conservativeness. We compare several strong baselines, and show that they are either more conservative than our algorithm, or do not satisfy the required risk bound. Our second experiment introduces more complexity to the autonomous driving scenario, and decreases the risk threshold to a realistically low level. We show that our algorithm still exhibits good performance, while several alternative algorithms become overly conservative. Finally, we present a real-world demonstration of our algorithm running on an autonomous class 8 truck (see Fig.~\ref{fig:page1} and video in Appx.~\ref{video}).

\section{Related Work}

In this section we review work on planning and control which aims to compute plans with safety guarantees in stochastic environments, involving uncertainty over both the robot dynamics and the movement of dynamic agents. 
The most general formulation of this problem is a chance constrained partially observable Markov decision process~\cite{santana2016} (CC-POMDP), in which feasible solutions guarantee the satisfaction of risk constraints, and optimal solutions are maximally non-conservative. Optimally solving CC-POMDPs is NP-hard~\cite{khonji2019}, and thus efficient approximations are required for real-world applications. Several applications of CC-POMDPs to autonomous driving have been proposed~\cite{huang2018,li2019}, which use RHC to approximate the optimal policy.


Stochastic model predictive control (MPC)~\cite{mesbah2016,carvalho2014,ono2012} is an RHC approach which extends the applicability of classic MPC to achieve safe, but suboptimal planning in CC-POMDPs (and related formulations).
Other MPC approaches, such as robust MPC~\cite{bemporad1999}, tube MPC~\cite{Garimella-RSS-18}, and reachability-based trajectory design for dynamic environments (RTD-D)~\cite{vaskov_towards_2019} assume a set-based representation of uncertainty, rather than the probabilistic representation used by our work, by stochastic MPC, and by other chance-constrained approaches (e.g.~\cite{aoude_probabilistically_2013,ono_chance-constrained_2015}). Set-based predictions of dynamic obstacles can either be learned~\cite{vaskov_towards_2019,jacobs2019}, or based on worst-case assumptions like backward reachability~\cite{leung2020} (at the cost of greater conservatism).
Hybrid set-based and probabilistic approaches have also been proposed, for example combining the FaSTrack method~\cite{herbert2017}, which uses worst-case assumptions to bound the error dynamics of a closed-loop tracking controller, with a probabilistic approach to predicting pedestrian trajectories~\cite{fisac2018}. 
Our method models the dynamics of the system probabilistically, and uses probabilistic predictions of other agents' motions  (e.g.~\cite{aoude_probabilistically_2013,tang2019,cui2019}). This allows us to reduce conservatism by leveraging prior knowledge of agents' environment-dependent behavior.
 
Like our approach, RTD-D maintains a contingency plan which is guaranteed to safely bring the robot to a safe set if replanning fails. However, in RTD-D this contingency plan occurs at the \textit{end} of the planning horizon, whereas in our algorithm, it can be selected after one step into the planning horizon, once the first observation has been obtained (this is similar to the ``committed plan'' concept proposed by \cite{tordesillas2019}). Our approach has two benefits over RTD-D. First, it allows our algorithm to plan for low-latency closed-loop reactions, reducing conservatism. Second, it allows our algorithm to use a longer planning horizon, which can be matched to the effective prediction horizon. Both RTD-D and our examples in this paper use the concept of~\textit{passive safety}~\cite{macek2009}; more complex schemes may be possible~\cite{shalevshwartz2018,wabersich2019}, but for concreteness we do not explore them further here.

Another class of techniques for reducing conservatism, 
which attempt to emulate the behavior of closed-loop policies, 
predict using ``closed-loop'' heuristics to account for future observations~\cite{du_toit_robot_2012,yan_incorporating_2005}. In partially closed-loop (PCL) RHC~\cite{du_toit_robot_2012}, future belief updates are modeled by  conditioning the future belief state on the most likely future observation sequence. Technically, this is an optimistic approach, which can lead to violation of safety constraints. In this paper we compare our approach with a PCL-RHC baseline, and we apply PCL techniques within our algorithm to show that our framework can bound risk while using approximate belief updating heuristics to improve performance.




\section{Problem Statement}
\label{ps}

Consider a stochastic discrete-time time-invariant system,
\begin{equation}
x_{i+1} = f(x_i, u_i, w_i)
\end{equation}
\begin{equation}
y_{i} = h(x_i, v_i),
\end{equation}
where $x_{i} \in \mathbb{X} \subseteq \mathbb{R}^{n_x}$ is the state at step $i$; $u_{i} \in \mathbb{U} \subseteq \mathbb{R}^{n_u}$ is the control taken at step $i$; and $w_{i} \in \mathbb{W} \subseteq \mathbb{R}^{n_w}$ represents the uncertainty in the system dynamics model and is sampled from a known probability distribution $P_{w_i}$.
The system is partially observable, where $y_{i} \in \mathbb{Y} \subseteq \mathbb{R}^{n_y}$ is the observation at step $i$; and $v_{i} \in \mathbb{V} \subseteq \mathbb{R}^{n_v}$ is the measurement noise and is sampled from a known probability distribution $P_{v_i}$. The state transition function $f$ and the observation function $h$ are both deterministic functions.

For a finite horizon $T$, the step-wise cost function $l(x_i, u_i)$ for each step $i$ and the final cost function $l_f(x_T)$ are defined based on the objective of the planning problem. 
In a finite horizon POMDP, the planning problem is to compute a non-stationary control policy $\Pi = \{\pi_0(\cdot),\cdots,\pi_{T-1}(\cdot)\}$, where $\pi_i(\cdot)$ maps the 
belief state at step $i$, $b_i \in \mathbb{B}$ to a control $u_i$ which minimizes the expected accumulated cost. The belief state is a sufficient statistic for the initial belief $b_0$ and the history of controls $u_{0:i-1}$ and  observations $y_{0:i}$.
The belief state transition function $b_{i+1}(x_{i+1}) = f_b(b_i, u_i, y_{i+1})$ updates the belief after control $u_i$ and observation $y_{i+1}$ by applying Bayesian filtering:
\begin{equation}
b_{i+1}(x_{i+1}) = \eta p(y_{i+1}|x_{i+1})\sum_{x_i\in\mathbb{X}}{p(x_{i+1}|x_i, u_i)b_i(x_i)}.
\end{equation}

For safety-critical applications, the states may be constrained to be outside some collision zone $\mathbb{X}^{coll} \subseteq \mathbb{X}$.
In this paper, we exclude states which are \textit{passively safe}~\cite{macek2009} from $\mathbb{X}^{coll}$: if the ego vehicle is static in state $x$, then $x \notin \mathbb{X}^{coll}$ and it is passively safe.
Since the uncertainty introduced to the system can be unbounded, it cannot be guaranteed that the state constraint is always satisfied. Instead, we apply chance constraints in the system, which bound the allowable probability of collision over one or more steps.

The joint chance-constrained POMDP optimization problem is formulated as:
\begin{subequations}
\label{eqn:p2}
\begin{equation}
\label{eqn:p2_opt}
\Pi^* = \arg\min_{\Pi}\mathop{\mathbb{E}}_{y_{0:T-1}}
\big[c_f(b_T) + \sum_{i=0}^{T-1}{c(b_i, \pi_i(b_i))}\big]
\end{equation}
such that:
\begin{align}
&b_{i+1} = f_b(b_i, \pi_i(b_i), y_{i+1}) 
\\
\label{eqn:p2_cc}
&\Pr\left( \bigcup\limits_{i = 0}^{T} (x_i \in \mathbb{X}^{coll}) \right) \leq \alpha,
\end{align}
\end{subequations}
where $c(b_i, \pi_i(b_i)) := \sum_{x_i\in\mathbb{X}}{l(x_i, \pi_i(b_i))b_i(x_i)}$ and $c_f(b_T) := \sum_{x_T\in\mathbb{X}}{l_f(x_T)b_T(x_T)}$ are the step-wise cost function and the final cost function in terms of the belief state, respectively.
The joint chance constraint \eqref{eqn:p2_cc} sets an upper bound of $\alpha$ on the total collision probability over $T$ steps,
potentially reducing conservatism by allowing non-uniform risk allocation over the planning horizon~\cite{ono_chance-constrained_2015}.

\section{Receding Horizon Control}
\label{rhc}

The optimization in \eqref{eqn:p2} is NP-hard~\cite{khonji2019}, and thus efficient approximations are needed for real-time applications. A well-established approach is to use an RHC strategy (e.g.~\cite{huang2018}), recursively solving \eqref{eqn:p2} from the current belief over a much shorter horizon $N$ than the original horizon $T$, executing the first control using the optimized policy, and updating the belief. When constraints of the form \eqref{eqn:p2_cc} are incorporated this is known as joint chance-constrained RHC (JCC-RHC).

\subsubsection*{Worst-Case Risk}

Although the JCC-RHC policy is suboptimal, intuitively, we might expect that by setting the joint chance constraint on every iteration equal to $\alpha N / T$, which on average allocates $\alpha / T$ for every step in the RHC planning horizon $N$, JCC-RHC can satisfy the original joint chance constraint of $\alpha$ over the interval $[0, T]$ in \eqref{eqn:p2_cc}.
However, we now sketch a proof that this does not hold (the full proof is in Appx. \ref{counter_example}).
The proof shows that at every iteration of RHC, JCC-RHC can in general allocate the maximum risk allowed by the joint chance constraint, $\alpha N / T$, to the first step of the plan. Over $T$ steps, the algorithm can incur risk equal to $\alpha N$, exceeding the desired joint chance constraint by a factor of $N$. This negates the benefit of joint chance constraints, and motivates the risk-budgeting framework we propose in Section~\ref{method}.

\subsubsection*{Open-loop Approximation}

Even with a short planning horizon $N$, solving the CC-POMDP may be too expensive to perform online. MPC techniques improve computational efficiency by transforming the CC-POMDP to an open-loop system, then applying RHC. The UMDP transformation (Appx.~\ref{umdp})~\cite{hauskrecht2000} makes the simple assumption that the POMDP is unobservable, with the  future belief state $b^{OL}$ fully determined by the control sequence using the open-loop belief state transition function $b^{OL}_{i+1}(x_{i+1}) = f^{OL}_b(b^{OL}_i, u_i)$:
\begin{equation}
\label{eqn:p3_ol}
b^{OL}_{i+1}(x_{i+1}) = \sum_{x_i\in\mathbb{X}}{p(x_{i+1}|x_i, u_i)b^{OL}_i(x_i)}.
\end{equation}

The UMDP solution satisfies the chance constraints of the original CC-POMDP, but it is a conservative approximation because the predicted open-loop belief states can have high covariance (e.g., imagine navigating a dynamic environment with closed eyes).
In contrast to UMDP, the partially closed-loop (PCL) transformation~\cite{du_toit_robot_2012} to an open-loop system is optimistic, but sacrifices safety guarantees. Next, in Section~\ref{method} we show that our algorithm achieves bounded risk, but is still compatible with approximate belief updating heuristics, like PCL, which are less conservative than UMDP.
\section{Risk-Budget RHC}
\label{method}
In this section, we first define a space of parametric bounds on the total risk incurred by a planning algorithm over a given episode or time interval.
We then introduce our safe and non-conservative planning algorithm, Risk-Budget RHC (RB-RHC), and prove a bound on the risk it incurs as a function of the algorithmic parameters.

\subsection{Interval Risk Bounds}

Interval risk bounds represent a natural, flexible criterion that can be used to constrain the risk incurred by a planner over arbitrary time periods or over the duration of episodic tasks, as well as to constrain the average rate at which risk is incurred over time. 

\textit{Definition 1:} An \textbf{interval risk bound} (IRB) is an upper bound on the risk incurred during any subinterval $[0, T]$ of the given time period or episodic task such that:
\begin{equation}
\label{eqn:irb}
    \Pr\left( \bigcup\limits_{i = 0}^{T} (x_i \in \mathbb{X}^{coll}) \right) \leq
    \rho_0 + \delta T,
\end{equation}
where $\rho_0$ and $\delta$ are parameters in $\mathbb{R}^+ \cup \{0\}$. The parameter $\rho_0$ captures the fixed component of the risk bound, while $\delta$ determines how rapidly the risk bound increases with time.
For example, system designers may wish to bound the probability $p$ of safety violation per $10^6$ steps ($\rho_0{=}0, \delta{=}p/10^6$), or to bound the safety violation probability $q$ while executing a particular task ($\rho_0{=}q, \delta{=}0$).

\subsection{Algorithm}
\label{algorithm}

We now define
Risk Budget RHC (RB-RHC), our algorithm for safe, non-conservative planning.
The key idea of RB-RHC is to maintain a dynamic risk budget
which constrains the amount of risk that each iteration of RB-RHC can allocate over the planning horizon in order to satisfy the desired IRB. 
Given an IRB, the risk budget, denoted $\rho_k$, is initialized at $k{=}0$ to $\rho_0$, corresponding to the fixed component of the IRB defined in \eqref{eqn:irb}. An additional $\delta$ is added to the risk budget at every iteration, corresponding to the component of the IRB in \eqref{eqn:irb} that increases over time.
To enforce the IRB using the risk budget, we first apply Boole's inequality to bound the LHS of \eqref{eqn:irb} by the sum of the risk incurred over all steps~\cite{4739221}.
At each iteration of RB-RHC, we decrease the risk budget by the amount of risk incurred by the action taken at that iteration, to ensure that the sum of the risk allowed over all steps cannot exceed the IRB.

At each iteration of RHC, if the risk budget becomes insufficient to plan, a contingency plan which is guaranteed to reach a safe set without exceeding the risk budget is executed immediately.
A natural definition for the contingency plan in the passive safety setting is the emergency stop action. 
Emergency stop is defined as executing the maximum deceleration control $u_{stop}$ for at most $t_{stop}$ steps,\footnote{$u_{stop}$ and $t_{stop}$ are constants which depend on the maximum deceleration and maximum speed of the vehicle.} which is assumed to reach the set of ``stopped'' belief states $\mathbb{B}^{stop} \subseteq \mathbb{B}$ from any initial belief state. 
Let $g_b(b) := \sum_{x \in \mathbb{X}^{coll}}b(x)$ be the probability of collision in belief state $b$. Belief states in $\mathbb{B}^{stop}$ have deterministic zero velocity; since we define passively safe states to be outside of $\mathbb{X}^{coll}$, the collision probability for belief states in $\mathbb{B}^{stop}$ is zero: $\forall b \in \mathbb{B}^{stop}: g_b(b) = 0.$
To bound the probability of collision during an emergency stop, we first define the function $f^{stop}_b(b, \tau)$, which returns the belief state after taking $u_{stop}$ for $\tau$ steps from a belief state $b$ without any observation received: $f^{stop}_b(b, \tau)= f^{OL}_b(f^{stop}_b(b, \tau{-}1), u_{stop})$.
We can then define $g^{stop}_b(b) := \sum^{t_{stop}}_{\tau=1}g_b(f^{stop}_b(b, \tau))$, which is an upper bound on the probability of collision during the emergency stop starting from a belief state $b$. 

At each step, RB-RHC solves the joint chance-constrained optimization problem:
\begin{subequations}
\label{eqn:p4}
\begin{equation}
\label{eqn:p4_opt}
u_{k:k+N-1}^* = \arg\min_{u_{k:k+N-1}}
(c_f(b_{k+N}) + \sum_{i=k}^{k+N-1}{c(b_i, u_i)})
\end{equation}
such that:
\begin{align}
\label{eqn:p4_belief}
&b_{k+1} = f^{OL}_b(b_k, u_k) \\
\label{eqn:p4_pcl}
&b_{i+1} = f^{\sim}_b(b_i, u_i), \forall i \in \{k{+}1, \ldots,k{+}N{-}1\} \\
\label{eqn:p4_cc}
&\sum^{k+N}_{i=k+1}\big[g_b(b_i) + g^{stop}_b(b_i)\big] \leq \rho_k.
\end{align}
\end{subequations}
The open-loop belief updating from step $k$ to $k{+1}$ in \eqref{eqn:p4_belief}, and within $g_b^{stop}$ in \eqref{eqn:p4_cc} ensure that if \eqref{eqn:p4} is solvable at step $k$, then we can execute $u^*_k$ at step $k$, followed by $u_{stop}$ (the contingency plan) at step $k{+1}$, incurring at most $\rho_k$ risk. 
The belief update notation $f^{\sim}_b$ in \eqref{eqn:p4_pcl} denotes that our algorithm can use approximate belief updating heuristics after step $k{+}1$ (such as partially closed-loop belief updating, which we use in our experiments, or sampling) to reduce the conservatism of purely open-loop belief updating, without sacrificing the safety guarantees that open-loop updating provides.
The risk usage for future steps ($i{>}k{+}1$) is estimated using the heuristic belief updates as $g_b(b^{~}_i) + g^{stop}_b(b^{~}_i)$ in \eqref{eqn:p4_cc}, which allows the algorithm to plan for the future risk allocation to remain within the risk budget $\rho_k$ at each iteration.

RB-RHC is defined in Algorithm \ref{alg2}. First, line \ref{alg2_sol} attempts to solve the open-loop planning problem \eqref{eqn:p4}. If \eqref{eqn:p4} does not have a solution, the ego vehicle must already be in a dangerous belief state. Thus, on line \ref{alg2_uk_choice}, we either choose $u_{stop}$ or $\text{NO--OP}$ (remain stopped) depending on whether the ego vehicle is moving or not. After the action is executed, the risk budget is increased by $\delta$ in line \ref{alg2_inc}.

A critical step of RB-RHC lies in line \ref{alg2_upd_rb}: whenever a new control $u^*_k \notin \{u_{stop}, $\text{ NO--OP}$\}$ is executed, the risk budget $\rho$ is decreased by subtracting out the risk incurred by $u^*_k$: $g_b(f^{OL}_b(b, u^*_k)) + g^{stop}_b(f^{OL}_b(b, u^*_k))$.
These terms completely capture the two possible causes of collision risk at the next step $k{+}1$ after $u^*_k$ is executed: the risk of collision following the open loop belief update ($g_b(f^{OL}_b(b, u^*_k))$), and the risk of collision during emergency stop if \eqref{eqn:p4} does not have a solution at step $k{+}1$ after $u^*_k$ is executed ($g^{stop}_b(f^{OL}_b(b, u^*_k))$). Subtracting the risk used from the budget at every step ensures that we can never exceed the total amount allocated to the budget, which is $\rho_0 + \delta T$.

\begin{algorithm}
\SetAlgoLined
\SetKwInOut{Input}{Input}
\SetKwFor{RepTimes}{repeat}{times}{end}
\Input{$b_0$, $\rho_0$, $\delta$, Maximum Time Horizon $T$}
$b \leftarrow b_0$; $\rho \leftarrow \rho_0$\;
\RepTimes{T}{
Solve (\ref{eqn:p4}) with $b_k=b$, $\rho_k=\rho$ for $u^*_k$\;\label{alg2_sol}
\If{(\ref{eqn:p4}) has solution with $b_k=b$, $\rho_k=\rho$\label{alg2_has_sol}}{
  $\rho \leftarrow \rho - g_b(f^{OL}_b(b, u^*_k)) - g^{stop}_b(f^{OL}_b(b, u^*_k))$\;\label{alg2_upd_rb}
}
\Else{
$
u^*_k \leftarrow 
\begin{cases}
u_{stop} & \text{ if } b \notin \mathbb{B}^{stop}\\
\text{NO--OP} & \text{ otherwise} 
\end{cases}
$ \label{alg2_uk_choice}
}
Execute $u^*_k$; Receive new observation $y_k$\;
$\rho \leftarrow \rho + \delta$\;\label{alg2_inc}
$b \leftarrow f_b(b, u^*_k, y_k)$\;
}
\caption{RB-RHC}
\label{alg2}
\end{algorithm}

\subsection{Theoretical Analysis of Safety Guarantee}
\label{proof}
Assume that the initial belief state $b_0$ satisfies the condition that (\ref{eqn:p4}) has a feasible solution at $k=0$ so we start with a non-irrecoverable state. We first prove by induction that the collision chance in the interval $[k, T]$ is bounded by $\rho_k + \delta (T-k)$ in RB-RHC when (\ref{eqn:p4}) is solvable at $k$. When $k=0$, RB-RHC satisfies the IRB $\rho_0 + \delta T$. 

\begin{lemma}
\label{lemma:induction}
Given the hypothesis below, the collision chance in $[k, T]$ is bounded by $\rho_k + \delta (T-k)$ in RB-RHC when (\ref{eqn:p4}) has a feasible solution at $k$. Hypothesis: For all possible future observations after $u^*_k$ is executed, at the first step $l>k$ when (\ref{eqn:p4}) is solvable again, the collision chance in $[l, T]$ is bounded by $\rho_l + \delta (T-l)$.

\proof We first show that $\rho_l + \delta (T-l) = \rho^\prime_k + \delta (T-k)$ is independent of $l$, where $\rho^\prime_k = \rho_k - g_b(f^{OL}_b(b_k, u^*_k)) - g^{stop}_b(f^{OL}_b(b_k, u^*_k))$ is the risk budget after the risk incurred by $u^*_k$ is subtracted. The reason is that in $[k, l]$, after the risk incurred by $u^*_k$ is subtracted from $\rho_k$, the risk budget will simply increase by $\delta$ at every step so $\rho_l=\rho^\prime_k + \delta (l-k)$.

We now calculate the collision chance in the interval $[k, T]$, which depends on the following three terms:

\noindent 1) Collision chance in $[k, k+1]$: $\Pr(x_{k+1} \in \mathbb{X}^{coll}) = g_b(f^{OL}_b(b_k, u^*_k)).$

\noindent 2) Collision chance in $[k+1, T]$ when (\ref{eqn:p4}) is solvable at $k+1$:
With the hypothesis when $l=k+1$,
$\Pr(b_{k+1}\in \mathbb{B}^{solv}_{k+1}, \bigcup\limits_{i = k+2}^{T} (x_i \in \mathbb{X}^{coll})) 
\le (\rho^\prime_k + \delta(T-k))\cdot\Pr(b_{k+1}\in \mathbb{B}^{solv}_{k+1}),$
where $\mathbb{B}^{solv}_{k+1}\subseteq \mathbb{B}$ denotes the set of $b_{k+1}$ such that \eqref{eqn:p4} is solvable at $k+1$.

\noindent 3) Collision chance in $[k+1, T]$ when (\ref{eqn:p4}) is unsolvable at $k+1$: RB-RHC will first execute a sequence of $u_{stop}$ followed by a sequence of $\text{NO--OP}$ until (\ref{eqn:p4}) is solvable again at step $l$. The collision chance is the sum of a) collision chance in $[k+1, l]$, which is bounded by $g^{stop}_b(f^{OL}_b(b_k, u^*_k))$ and b) collision chance in $[l, T]$, which is bounded by $(\rho^\prime_k + \delta(T-k))\cdot\Pr(b_{k+1}\notin \mathbb{B}^{solv}_{k+1})$ using the hypothesis.

By summing the collision chance of the three conditions above, we prove the collision chance in $[k,T]$ is bounded by $\rho_k + \delta(T-k)$:
\begin{equation*}
\begin{split}
&g_b(f^{OL}_b(b_k, u^*_k)) + (\rho^\prime_k + \delta(T-k))\cdot\Pr(b_{k+1}\in \mathbb{B}^{solv}_{k+1}) \\ 
&+ (\rho^\prime_k + \delta(T-k))\cdot\Pr(b_{k+1}\notin \mathbb{B}^{solv}_{k+1}) + g^{stop}_b(f^{OL}_b(b_k, u^*_k)) \\
&= g_b(f^{OL}_b(b_k, u^*_k)) + (\rho^\prime_k + \delta(T-k)) + g^{stop}_b(f^{OL}_b(b_k, u^*_k)) \\
&= (\rho_k + \delta(T-k)).
\end{split}
\end{equation*}
\endproof
\end{lemma}

\begin{theorem}
RB-RHC satisfies IRB with $\rho_0 + \delta T$ when starting with a non-irrecoverable state.

\proof
With Lemma \ref{lemma:induction} and induction at $k=0$, the theorem is proven.
\endproof
\end{theorem}

\section{Simulation and Real-World Results}
\label{results}
We demonstrate the safety and non-conservativeness of RB-RHC in two simulated scenarios and in a mixed-reality scenario using a class 8 autonomous truck. 
We compare our algorithm with three joint chance-constrained baselines: JCC-FH uses UMDP to do finite-horizon open-loop planning with no replanning; JCC-RHC uses UMDP to plan with a receding horizon; and PCL-RHC uses a partially closed-loop belief updating heuristic~\cite{du_toit_robot_2012} to plan non-conservatively.


\subsubsection*{Implementation Details}

We assume that a kinematically-feasible reference path which is collision-free with respect to all static obstacles is provided by a path planner.
RB-RHC and all baselines plan a speed profile over the reference path to reach the goal with minimum cost, while satisfying a bound on the probability of collision with other vehicles.
The collision probability between two vehicles is computed by covering the vehicle footprints by disks and summing the collision probabilities between each pair of disks, which can be evaluated in closed form.
The ego dynamics are
governed by the vehicle kinematic model and a model of the controller tracking error, assumed to be Gaussian distributed. 
The world state represents the state of the ego vehicle and all other vehicles, and is assumed to be known at the current timestep, apart from other vehicles' intended motion patterns. Other vehicles' motion patterns are modeled as a mixture of RR-GP distributions~\cite{aoude_probabilistically_2013}, assumed to be independent from the ego vehicle state. For each motion pattern, RB-RHC uses PCL for the belief update $f_b^{\sim}$ in \eqref{eqn:p4_pcl}.
Problem (\ref{eqn:p4}) is solved by graph search in belief space. 
%
%
To address the chance constraint (\ref{eqn:p4_cc}), we introduce the Lagrange multiplier $\lambda$, and
use a bisection search method to find a value of $\lambda$ which corresponds to a locally optimal solution to \eqref{eqn:p4}~\cite{ono_chance-constrained_2015}. (Read more in Appx.~\ref{extended_details}).


\subsubsection*{Experiment 1}
\begin{figure}[hbt]
\centering
\includegraphics[height=3.4cm]{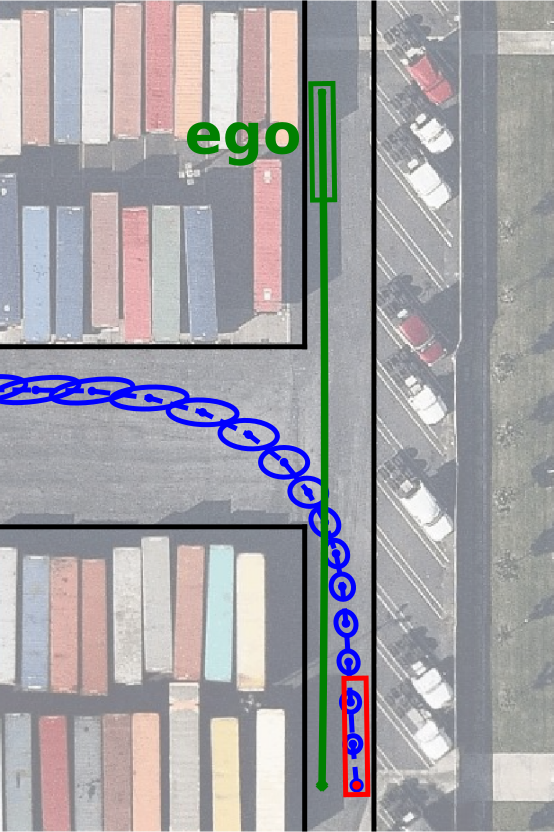}
\quad
\includegraphics[height=3.4cm]{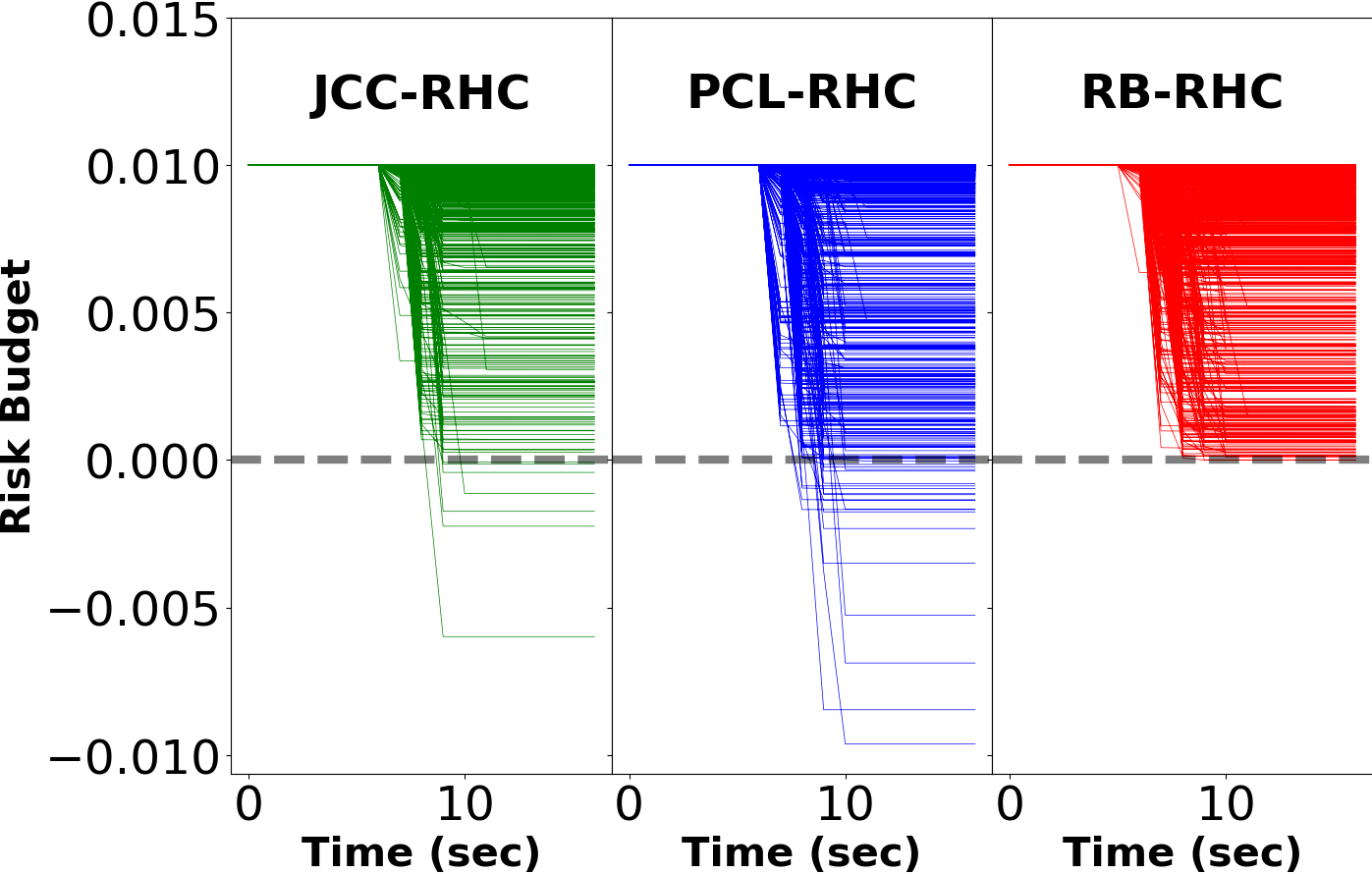}
\caption{Experiment 1. \textit{Left:} Scenario setup: ego starting pose (green), dynamic obstacle starting pose (red), position distribution of the dynamic obstacle (blue), ego path (green line).
\textit{Right:} Risk budget over time. Only RB-RHC never exceeds the risk budget.}
\label{fig:exp1}
\end{figure}

This experiment shows that RB-RHC satisfies the required risk bound, while PCL-RHC violates it, in a scenario in which the ego vehicle interacts with another vehicle turning left at a T-junction in a trailer yard environment (Fig.~\ref{fig:exp1}, left). For simplicity, both vehicles have a bus-like $12.6\,\mathrm{m} \times 2.4\,\mathrm{m}$ rectangular shape, with kinematics given by the deterministic bicycle model. We assume the other vehicle follows a single motion pattern. The current state is fully observable; however, the ego vehicle is uncertain about the future states of itself and the other vehicle.

The desired IRB has $\rho_0 = 0.01, \delta=0$.\footnote{We use a large value of $\rho_0=0.01$, because it allows us to empirically measure collision probability in a reasonable number of experimental trials.}
The maximum planning horizon is $T = N = 25 \,\text{sec}$, the planning timestep is $1\, \text{sec}$ and the re-planning frequency is $1\, \mathrm{Hz}$.
To train the RR-GP model of the other vehicle, we collected a small dataset of manually driven left turn trajectories.

Table \ref{tab:exp1} presents averaged results of each algorithm over $1000$ sampled trials (sampling the initial state of the ego vehicle and the trajectory of the other vehicle from the RR-GP prediction model). Only PCL-RHC violates the risk bound, while RB-RHC  satisfies the bound statistically. The fact that the failure rate in RB-RHC exactly matches $\rho_0$ suggests that RB-RHC takes the maximum amount of risk allowed in order to minimize the incurred cost. The collision rate for JCC-RHC is lower than $\rho_0$ because in practice it is often conservative. In theory, JCC-RHC can exceed the desired risk bound when multiple risky events are introduced, as shown in Section \ref{rhc}, but this is not observed in this scenario, because it contains only one risky event. Additionally, in Fig.~\ref{fig:exp1} (right), we can see that both JCC-RHC and PCL-RHC occasionally exceed the risk budget $\rho_0$, while RB-RHC always remains within the budget.

\begin{table}[t]
\vspace{7px}
\caption{Simulation results, Experiment $1$, $\rho_0 = 0.01, \delta=0$}
\label{tab:exp1}
\begin{center}
\begin{tabular}{|c||c|c|}
\hline
Algorithm & Collision Rate & Navigation Cost\\
\hline
JCC-FH & 0.007 $\pm$ 0.003 & 24.678 $\pm$ 0.037\\
\hline
JCC-RHC & 0.001 $\pm$ 0.001 & 24.177 $\pm$ 0.048\\
\hline
PCL-RHC & 0.015 $\pm$ 0.004 & 23.866 $\pm$ 0.086\\
\hline
RB-RHC & 0.010 $\pm$ 0.003 & 24.021 $\pm$ 0.081\\
\hline
\end{tabular}
\end{center}
\end{table}

\subsubsection*{Experiment 2}
\begin{figure}[hb]
\centering
\includegraphics[height=3.4cm]{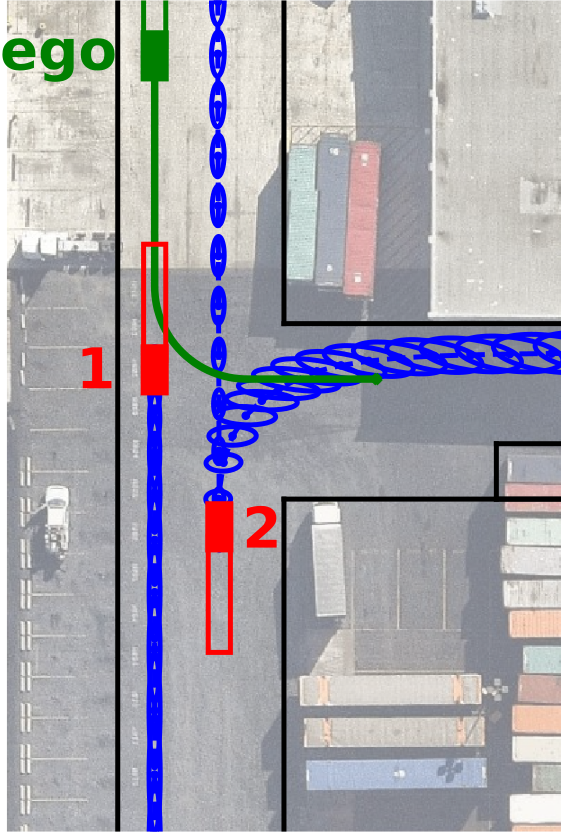}
\quad
\includegraphics[height=3.4cm]{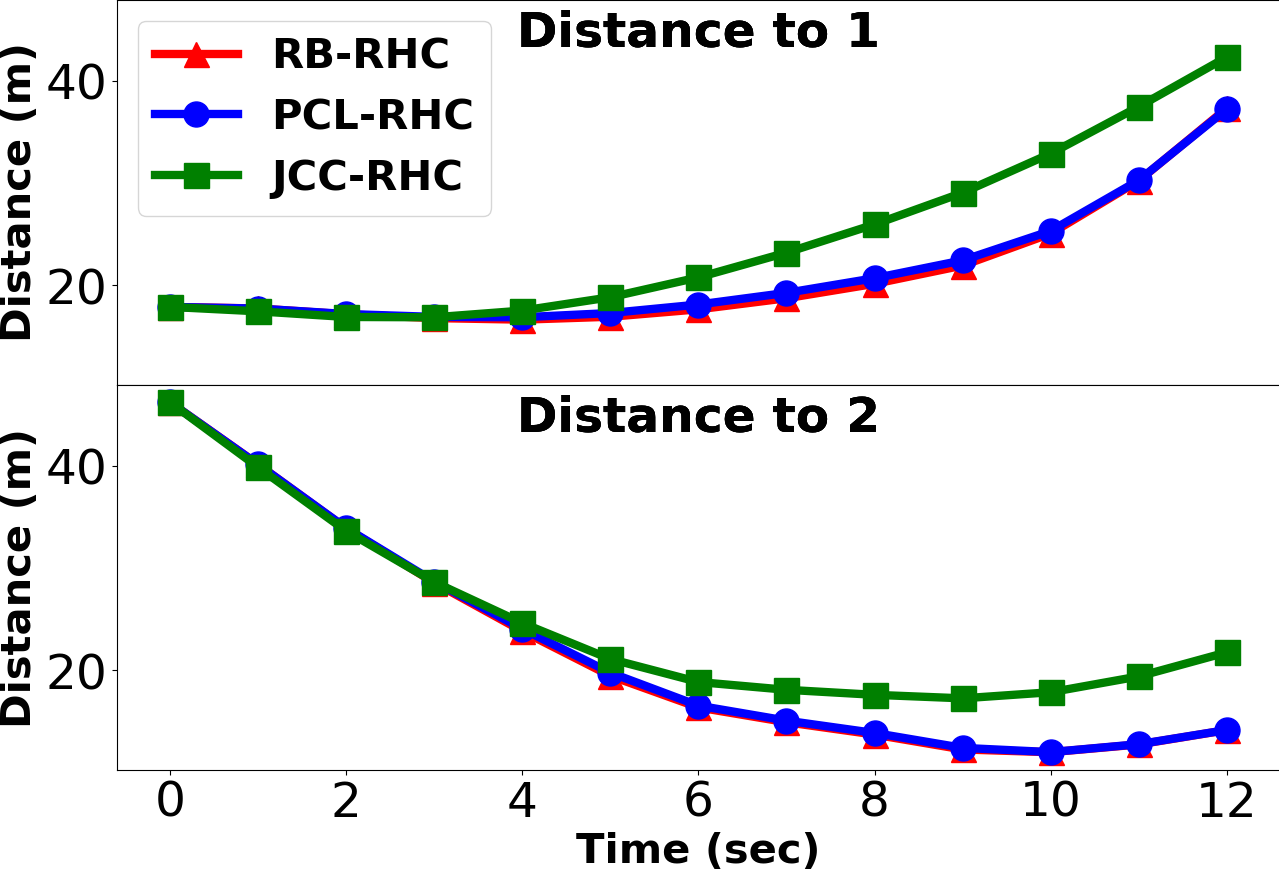}
\caption{Experiment 2. \textit{Left:} Scenario setup: ego starting pose (green), dynamic obstacles' starting pose (red; tractor filled, trailer unfilled), position distribution of the dynamic obstacles (blue), ego path (green line).
\textit{Right:} Average distance between ego vehicle and other vehicles over time.}
\label{fig:exp2}
\end{figure}

This experiment shows that RB-RHC is less conservative than JCC-RHC
in a complex scenario involving three interacting vehicles. The ego objective is to make a left turn while avoiding collision with vehicle $1$ moving ahead of the ego, and vehicle $2$ coming from the opposite direction with an uncertain intention (Fig.~\ref{fig:exp2}, left). We have a prior that vehicle $2$ will turn right with $70\%$ probability or go straight with $30\%$ probability, but its intention is not directly observable. Each vehicle is a $5.0\,\mathrm{m} \times 2.5\,\mathrm{m}$ tractor with a $12.5\,\mathrm{m}\times2.4\,\mathrm{m}$ trailer, with kinematics following a deterministic tractor-trailer model.

The required IRB has $\rho_0=10^{-6}, \delta=0$, which roughly corresponds to one expected collision per year of vehicle operation time, on par with the collision rate of human drivers in trailer yards. The planning horizon, planning timestep, and re-planning frequency
are the same as in Experiment $1$. We again train the RR-GP prediction model for all three motion patterns with a small, manually collected dataset. 

Table \ref{tab:exp2} presents averaged results of each algorithm over $1000$ sampled trials. Due to the small value of $\rho_0$, we did not observe any collisions in any of the $1000$ simulation trials. The cost of RB-RHC is $9.6\%$ lower than the cost of JCC-RHC, demonstrating that RB-RHC is less conservative than JCC-RHC. PCL-RHC and RB-RHC incur similar cost, and JCC-FH incurs more cost than any of the RHC-based algorithms.  Fig.~\ref{fig:exp2} (right) shows that while all compared algorithms maintain a safe distance to other vehicles on average, JCC-RHC conservatively maintains a larger distance to other vehicles than RB-RHC and PCL-RHC.

\subsubsection*{Real-World Trailer Yard Demonstrations}

\begin{figure}[htb]
    \centering
    \includegraphics[width=\linewidth]{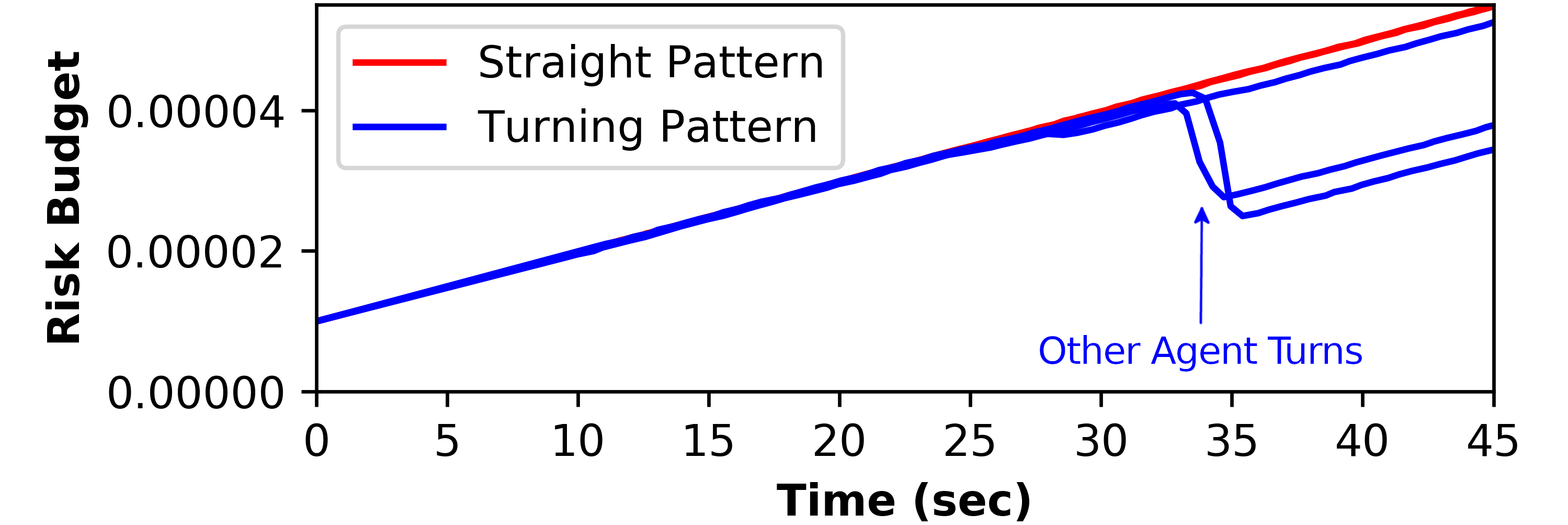}
    \caption{Real-world demonstration. Risk budget usage over time. Three trials for each motion pattern of the other vehicle are shown.}
    \label{fig:real}
\end{figure}

To demonstrate the applicability of RB-RHC in a real-world autonomous system, we deploy RB-RHC as a speed planning algorithm onboard an autonomous class 8 truck. The objective is to turn right into an aisle while avoiding an oncoming tractor-trailer, which may either go straight or make a left turn (see Fig.~\ref{fig:page1}, top). Static obstacles represent parked trailers, and dynamic obstacles represent other human-driven trucks operating in the trailer yard; for safety reasons these are all virtual obstacles simulated in a mixed-reality environment. In this experiment, the desired IRB has $\rho_0 = 10^{-5}, \delta = 10^{-6}$. Our RB-RHC planner is implemented in C++ and re-plans at $20$ Hz. The output speed profile is tracked by a low-level real-time controller.
Despite the potential for errors when modeling a real-world, 12 metric ton tractor-trailer system, the system is able to successfully maneuver the vehicle and remain within the risk budget. Fig.~\ref{fig:real} shows that when the dynamic obstacle follows the turning pattern, the risk budget is indeed used primarily at the turning point. The video in Appx.~\ref{video} shows that the truck maintains a safe but not overly conservative distance to the dynamic obstacle.

\begin{table}[t]
\vspace{5px}
\caption{Simulation results, Experiment $2$, $\rho_0 = 10^{-6}, \delta=0$}
\label{tab:exp2}
\begin{center}
\begin{tabular}{|c||c|c|}
\hline
Algorithm & Collision Rate & Navigation Cost\\
\hline
JCC-FH & 0.0 $\pm$ 0.0 & 29.153 $\pm$ 0.078\\
\hline
JCC-RHC & 0.0 $\pm$ 0.0 & 21.280 $\pm$ 0.075\\
\hline
PCL-RHC & 0.0 $\pm$ 0.0 & 19.301 $\pm$ 0.058\\
\hline
RB-RHC & 0.0 $\pm$ 0.0 & 19.246 $\pm$ 0.061\\
\hline
\end{tabular}
\end{center}
\end{table}

\vspace{-0.5cm}
\section{Conclusion}
In order to enable deployment of autonomous robots in safety-critical applications, the robotic system has to be at least as safe as a human, while at the same time maintaining near human-level efficiency at the task.
In this paper, we have shown that existing techniques for chance-constrained planning either violate the required safety level or exhibit overly conservative behavior. We presented a new framework for designing safe, non-conservative planning algorithms for dynamic uncertain environments. To quantify a system's level of safety, we defined interval risk bounds, which provide a parametric upper bound on the risk incurred by an algorithm during a given episode or time interval. We derived a novel RHC algorithm called Risk Budget RHC, proved that it can satisfy a desired IRB, and empirically demonstrated that our algorithm is both more safe and less conservative than several strong baselines in two experiments in simulated dynamic environments, and in a real-world demonstration of our algorithm running on an autonomous class 8 truck.

\bibliographystyle{IEEEtran}
\bibliography{reference}

\begin{thebibliography}{10}
\providecommand{\url}[1]{#1}
\csname url@samestyle\endcsname
\providecommand{\newblock}{\relax}
\providecommand{\bibinfo}[2]{#2}
\providecommand{\BIBentrySTDinterwordspacing}{\spaceskip=0pt\relax}
\providecommand{\BIBentryALTinterwordstretchfactor}{4}
\providecommand{\BIBentryALTinterwordspacing}{\spaceskip=\fontdimen2\font plus
\BIBentryALTinterwordstretchfactor\fontdimen3\font minus
  \fontdimen4\font\relax}
\providecommand{\BIBforeignlanguage}[2]{{%
\expandafter\ifx\csname l@#1\endcsname\relax
\typeout{** WARNING: IEEEtran.bst: No hyphenation pattern has been}%
\typeout{** loaded for the language `#1'. Using the pattern for}%
\typeout{** the default language instead.}%
\else
\language=\csname l@#1\endcsname
\fi
#2}}
\providecommand{\BIBdecl}{\relax}
\BIBdecl

\bibitem{tang2019}
Y.~C. Tang and R.~Salakhutdinov, ``Multiple futures prediction,'' in
  \emph{Advances in Neural Information Processing Systems}, 2019.

\bibitem{cui2019}
H.~Cui, V.~Radosavljevic, F.-C. Chou, T.-H. Lin, T.~Nguyen, T.-K. Huang,
  J.~Schneider, and N.~Djuric, ``Multimodal trajectory predictions for
  autonomous driving using deep convolutional networks,'' in \emph{{IEEE}
  International Conference on Robotics and Automation ({ICRA})}, 2019.

\bibitem{jacobs2019}
H.~O. {Jacobs}, O.~K. {Hughes}, M.~{Johnson-Roberson}, and R.~{Vasudevan},
  ``Real-time certified probabilistic pedestrian forecasting,'' \emph{IEEE
  Robotics and Automation Letters}, vol.~2, no.~4, pp. 2064--2071, 2017.

\bibitem{khonji2019}
M.~Khonji, A.~Jasour, and B.~Williams, ``Approximability of constant-horizon
  constrained {POMDP},'' in \emph{Proceedings of the Twenty-Eighth
  International Joint Conference on Artificial Intelligence ({IJCAI})}, 2019.

\bibitem{ono2012}
M.~{Ono}, ``Joint chance-constrained model predictive control with
  probabilistic resolvability,'' in \emph{2012 American Control Conference
  (ACC)}, 2012, pp. 435--441.

\bibitem{macek2009}
K.~Ma\^cek, D.~Vasquez, T.~Fraichard, and R.~Siegwart, ``Towards safe vehicle
  navigation in dynamic urban scenarios,'' \emph{Automatika}, vol.~50, no. 3-4,
  pp. 184--–194, 2009.

\bibitem{santana2016}
P.~Santana, S.~Thi\`ebaux, and B.~Williams, ``{RAO}*: {A}n algorithm for chance
  constrained {POMDP}s,'' in \emph{Proceedings of the {AAAI} Conference on
  Artificial Intelligence}, 2016.

\bibitem{huang2018}
X.~{Huang}, A.~{Jasour}, M.~{Deyo}, A.~{Hofmann}, and B.~C. {Williams},
  ``Hybrid risk-aware conditional planning with applications in autonomous
  vehicles,'' in \emph{2018 IEEE Conference on Decision and Control (CDC)},
  2018, pp. 3608--3614.

\bibitem{li2019}
\BIBentryALTinterwordspacing
N.~Li, A.~Girard, and I.~Kolmanovsky, ``{Stochastic Predictive Control for
  Partially Observable Markov Decision Processes With Time-Joint Chance
  Constraints and Application to Autonomous Vehicle Control},'' \emph{Journal
  of Dynamic Systems, Measurement, and Control}, vol. 141, no.~7, 03 2019,
  071007. [Online]. Available: \url{https://doi.org/10.1115/1.4043115}
\BIBentrySTDinterwordspacing

\bibitem{mesbah2016}
A.~Mesbah, ``Stochastic model predictive control: An overview and perspectives
  for future research,'' \emph{{IEEE} Control Systems}, vol.~36, no.~6, pp.
  30--–44, 2016.

\bibitem{carvalho2014}
A.~Carvalho, Y.~Gao, S.~Lef\`evre, and F.~Borrelli, ``Stochastic predictive
  control of autonomous vehicles in uncertain environments,'' in \emph{12th
  International Symposium on Advanced Vehicle Control}, 2014, pp. 712--719.

\bibitem{bemporad1999}
A.~Bemporad and M.~Morari, ``Robust model predictive control: a survey,'' in
  \emph{Robustness in identification and control. Lecture Notes in Control and
  Information Sciences}.\hskip 1em plus 0.5em minus 0.4em\relax London:
  Springer, 1999, vol. 245.

\bibitem{Garimella-RSS-18}
G.~Garimella, M.~Sheckells, J.~Moore, and M.~Kobilarov, ``Robust obstacle
  avoidance using tube {NMPC},'' in \emph{Proceedings of Robotics: Science and
  Systems}, Pittsburgh, Pennsylvania, June 2018.

\bibitem{vaskov_towards_2019}
\BIBentryALTinterwordspacing
S.~Vaskov, S.~Kousik, H.~Larson, F.~Bu, J.~Robert~Ward, S.~Worrall,
  M.~Johnson-Roberson, and R.~Vasudevan, ``\BIBforeignlanguage{en}{Towards
  {Provably} {Not}-{At}-{Fault} {Control} of {Autonomous} {Robots} in
  {Arbitrary} {Dynamic} {Environments}},'' in
  \emph{\BIBforeignlanguage{en}{Robotics: {Science} and {Systems} {XV}}}.\hskip
  1em plus 0.5em minus 0.4em\relax Robotics: Science and Systems Foundation,
  Jun. 2019. [Online]. Available:
  \url{http://www.roboticsproceedings.org/rss15/p51.pdf}
\BIBentrySTDinterwordspacing

\bibitem{aoude_probabilistically_2013}
\BIBentryALTinterwordspacing
G.~S. Aoude, B.~D. Luders, J.~M. Joseph, N.~Roy, and J.~P. How,
  ``\BIBforeignlanguage{en}{Probabilistically safe motion planning to avoid
  dynamic obstacles with uncertain motion patterns},''
  \emph{\BIBforeignlanguage{en}{Autonomous Robots}}, vol.~35, no.~1, pp.
  51--76, Jul. 2013. [Online]. Available:
  \url{http://link.springer.com/10.1007/s10514-013-9334-3}
\BIBentrySTDinterwordspacing

\bibitem{ono_chance-constrained_2015}
\BIBentryALTinterwordspacing
M.~Ono, M.~Pavone, Y.~Kuwata, and J.~Balaram,
  ``\BIBforeignlanguage{en}{Chance-constrained dynamic programming with
  application to risk-aware robotic space exploration},''
  \emph{\BIBforeignlanguage{en}{Autonomous Robots}}, vol.~39, no.~4, pp.
  555--571, Dec. 2015. [Online]. Available:
  \url{http://link.springer.com/10.1007/s10514-015-9467-7}
\BIBentrySTDinterwordspacing

\bibitem{leung2020}
K.~Leung, E.~Schmerling, M.~Zhang, M.~Chen, J.~Talbot, J.~C. Gerdes, and
  M.~Pavone, ``On infusing reachability-based safety assurance within planning
  frameworks for human–robot vehicle interactions,'' \emph{The International
  Journal of Robotics Research}, vol.~39, no. 10-11, pp. 1326--1345, 2020.

\bibitem{herbert2017}
S.~Herbert, M.~Chen, S.~Han, S.~Bansal, J.~F. Fisac, and C.~J. Tomlin,
  ``{FaSTrack}: a modular framework for fast and guaranteed safe motion
  planning,'' in \emph{{IEEE} Conference on Decision and Control (CDC)}, 2017.

\bibitem{fisac2018}
J.~F. Fisac, A.~Bajcsy, S.~Herbert, D.~Fridovich-Keil, S.~Wang, C.~J. Tomlin,
  and A.~D. Dragan, ``Probabilistically safe robot planning with
  confidence-based human predictions,'' in \emph{Robotics Science and Systems
  (RSS)}, 2018.

\bibitem{tordesillas2019}
J.~{Tordesillas}, B.~T. {Lopez}, and J.~P. {How}, ``Faster: Fast and safe
  trajectory planner for flights in unknown environments,'' in \emph{2019
  IEEE/RSJ International Conference on Intelligent Robots and Systems (IROS)},
  2019, pp. 1934--1940.

\bibitem{shalevshwartz2018}
\BIBentryALTinterwordspacing
S.~Shalev{-}Shwartz, S.~Shammah, and A.~Shashua, ``On a formal model of safe
  and scalable self-driving cars,'' \emph{CoRR}, vol. abs/1708.06374, 2017.
  [Online]. Available: \url{http://arxiv.org/abs/1708.06374}
\BIBentrySTDinterwordspacing

\bibitem{wabersich2019}
K.~P. {Wabersich}, L.~{Hewing}, A.~{Carron}, and M.~N. {Zeilinger},
  ``Probabilistic model predictive safety certification for learning-based
  control,'' \emph{IEEE Transactions on Automatic Control}, pp. 1--1, 2021.

\bibitem{du_toit_robot_2012}
\BIBentryALTinterwordspacing
N.~E. Du~Toit and J.~W. Burdick, ``\BIBforeignlanguage{en}{Robot {Motion}
  {Planning} in {Dynamic}, {Uncertain} {Environments}},''
  \emph{\BIBforeignlanguage{en}{IEEE Transactions on Robotics}}, vol.~28,
  no.~1, pp. 101--115, Feb. 2012. [Online]. Available:
  \url{http://ieeexplore.ieee.org/document/6024480/}
\BIBentrySTDinterwordspacing

\bibitem{yan_incorporating_2005}
\BIBentryALTinterwordspacing
J.~Yan and R.~R. Bitmead, ``\BIBforeignlanguage{en}{Incorporating state
  estimation into model predictive control and its application to network
  traffic control},'' \emph{\BIBforeignlanguage{en}{Automatica}}, vol.~41,
  no.~4, pp. 595--604, Apr. 2005. [Online]. Available:
  \url{https://linkinghub.elsevier.com/retrieve/pii/S0005109805000130}
\BIBentrySTDinterwordspacing

\bibitem{hauskrecht2000}
M.~Hauskrecht, ``Value-function approximations for partially observable
  {M}arkov decision processes,'' \emph{Journal of Artificial Intelligence
  Research}, vol.~13, pp. 33--94, 2000.

\bibitem{4739221}
{Masahiro Ono} and B.~C. {Williams}, ``Iterative risk allocation: A new
  approach to robust model predictive control with a joint chance constraint,''
  in \emph{2008 47th IEEE Conference on Decision and Control}, 2008, pp.
  3427--3432.

\end{thebibliography}

\appendix

\subsection{Supplementary Video}
\label{video}
The supplementary video can be found at \texttt{https://youtu.be/697L1eLL9iw}. 
Its most important content is the real-world demonstration of our algorithm running on an autonomous class 8 truck in a trailer yard environment. The video summarizes the paper, and also sketches a proof that the worst-case risk incurred by JCC-RHC in an interval exceeds the value of the chance constraint over that interval.

\subsection{Worst-case Risk of JCC-RHC}
\label{counter_example}

Given a desired IRB with $\rho_0 = 0, \delta = \alpha / T$, we now use a counterexample to prove that JCC-RHC with a planning horizon $N$, which sets the joint chance constraint on every iteration equal to $\alpha N / T$ (as described in Section~\ref{rhc} of the main text), can violate such an IRB. Another counterexample can be found in the Supplementary Video (Appx. \ref{video}).

\proof 
Imagine a racetrack with two consecutive sharp curves. For the first sharp curve, there is $10\%$ chance of crashing if driven at $100$ mph, and a $0\%$ chance of crashing if driven at $70$ mph; for the second curve, there is $10\%$ chance of crashing if driven at $90$ mph, and a $0\%$ chance of crashing if driven at $70$ mph. For simplicity, assume no other speeds can be chosen. We solve this using RHC with $N=2$, where $u_1$ and $u_2$ are the speeds taken at the first and second curves, and $y_1$ and $y_2$ are their respective outcomes, which can either be ``crash'' or ``safe''. Our goal is to plan a policy over $u_1$ and $u_2$ which drives as fast as possible while bounding the probability of crashing during the interval $[0, 2]$ below $\alpha=0.1$.

\begin{figure}[htbp]
  \centering
  \includegraphics[width=0.5\columnwidth]{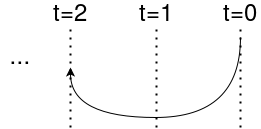}
  \caption{Illustration of the racetrack counterexample.}
\end{figure}

In order to realize this bound, we set the joint chance constraint to $\alpha N / T = 0.1 \cdot 2 / 2 = 0.1$, which allocates $\alpha / T$ risk for each step of the planning horizon. At $t=0$, the optimal policy is: $u^*_1 = 100, u^*_2=70$, which satisfies the joint chance constraint of $0.1$ by allocating all the risk to drive fast in the first curve. If the car luckily does not crash in the first curve, RHC can replan at $t=1$ using a joint chance constraint of $\alpha N / T= 0.1$ again and get $u^*_2=90$. The probability of all three outcome combinations with this RHC policy is:
\begin{equation*}
\begin{split}
&\Pr(y_1=\text{``crash''}) = 0.1 \\
&\Pr(y_1=\text{``safe''}, y_2=\text{``crash''}) = 0.9 \cdot 0.1 = 0.09 \\
&\Pr(y_1=\text{``safe''}, y_2=\text{``safe''}) = 0.9 \cdot 0.9 = 0.81.
\end{split}
\end{equation*}
The overall probability of crashing for this RHC policy is $0.1 + 0.09 = 0.19$. Therefore, RHC with a joint chance constraint which allocates $\alpha/T$ risk for each step of the planning horizon does not satisfy our desired chance constraint $\alpha = 0.1$, or the IRB with $\rho_0 = 0, \delta = \alpha / T$ on the interval $[0, T{=}2]$, which is $\delta T = \alpha =0.1$.
\endproof

\subsection{The UMDP Method}
\label{umdp}

The UMDP method solves the joint chance-constrained optimization problem:
\begin{subequations}
\label{eqn:p6}
\begin{equation}
\label{eqn:p6_opt}
u_{0:T-1}^* = \arg\min_{u_{0:T-1}}
(c_f(b_{T}) + \sum_{i=0}^{T-1}{c(b_i, u_i)})
\end{equation}
such that:
\begin{align}
\label{eqn:p6_ol}
&b_{i+1} = f^{OL}_b(b_i, u_i), \forall i \in \{0, \ldots,T{-}1\} \\
\label{eqn:p6_cc}
&\Pr\left( \bigcup\limits_{i = 0}^{T} (x_i \in \mathbb{X}^{coll}) \right) \leq \alpha,
\end{align}
\end{subequations}
and executes the optimal control sequence $u_{0:T-1}^{*}$ in an open-loop manner.
The optimization in (\ref{eqn:p6}) is an open-loop approximation of (\ref{eqn:p2}). Assuming no future observations will be available, the UMDP method computes a control sequence $u_{0:T-1}$ in (\ref{eqn:p6}) instead of a control policy $\Pi$ in (\ref{eqn:p2}). Since the open-loop belief update in (\ref{eqn:p6_ol}) is not a function of the observations, (\ref{eqn:p6_opt}) does not take the expectation over future observations $y_{0:T-1}$ as if in (\ref{eqn:p2_opt}), which greatly improves the computational efficiency.

In this part, we prove that the optimal control sequence $u_{0:T-1}^*$ computed using (\ref{eqn:p6}) satisfies the CC-POMDP chance constraint (\ref{eqn:p2_cc}). Expanding the LHS of (\ref{eqn:p2_cc}), we get:
\begin{equation*}
\begin{split}
&\Pr\left( \bigcup\limits_{i = 0}^{T} (x_i \in \mathbb{X}^{coll}) \right) \\ 
&=\int p(x_{0:T}, y_{0:T} | b_0, \Pi)\boldsymbol{\cdot} \mathbbm{1}(\bigcup\limits_{i = 0}^{T} (x_i \in \mathbb{X}^{coll})) \diff x_{0:T} \diff y_{0:T}.
\end{split}
\end{equation*}
In UMDP, $\Pi = u_{0:T-1}^*$, which is not a function of the belief; therefore, the equation above can be rewritten to:
\begin{alignat*}{2}
&\int &&p(x_0|b_0)p(y_0|x_0)\prod_{i=0}^{T-1}p(x_{i+1}|x_i, u_i^*)p(y_{i+1}|x_{i+1})\boldsymbol{\cdot}\\
& &&\mathbbm{1}(\bigcup\limits_{i = 0}^{T} (x_i \in \mathbb{X}^{coll})) \diff x_{0:T} \diff y_{0:T} \\
&=&&\underbrace{\int p(y_0|x_0)\prod_{i=0}^{T-1}p(y_{i+1}|x_{i+1})\diff y_{0:T}}_{=1}\boldsymbol{\cdot}\\
& &&\int p(x_{0:T} | b_0, u_{0:T-1}^*)\boldsymbol{\cdot} \mathbbm{1}(\bigcup\limits_{i = 0}^{T} (x_i \in \mathbb{X}^{coll})) \diff x_{0:T} \\
&=&&\int p(x_{0:T} | b_0, u_{0:T-1}^*)\boldsymbol{\cdot} \mathbbm{1}(\bigcup\limits_{i = 0}^{T} (x_i \in \mathbb{X}^{coll})) \diff x_{0:T},
\end{alignat*}
which is bounded by $\alpha$ because $u_{0:T-1}^*$ satisfies (\ref{eqn:p6_cc}).

For our JCC-FH implementation in Section \ref{results},  Boole's inequality is applied to bound the LHS of (\ref{eqn:p6_cc})~\cite{4739221} and replace the joint chance constraint condition (\ref{eqn:p6_cc}) with:
\begin{equation}
\label{eqn:p6_cc_new}
\sum^{T}_{i=0}\big[g_b(b_i)\big] \leq \alpha,
\end{equation}
where $g_b(b)$ is defined in Section \ref{algorithm} as the probability of collision in belief state $b$.

\subsection{Extended Implementation Details}
\label{extended_details}
This appendix contains an extended version of the Implementation Details paragraph in Section \ref{results}, which is condensed due to the space limitations of the conference format. We assume that a kinematically-feasible reference path which is collision-free with respect to all static obstacles is provided by a path planner.
RB-RHC and all baselines plan a speed profile over the reference path to reach the goal with minimum cost, while satisfying a bound on the probability of collision with other vehicles.
The collision probability between two vehicles is computed by covering the vehicle footprints by disks and summing the collision probabilities between each pair of disks, which can be upper bounded by a closed form function:
\begin{equation*}
0.5 + 0.5\cdot\erf(-d / 
\sqrt{2(\mu_{x_1} - \mu_{x_2})^\intercal(\Sigma_1 + \Sigma_2)(\mu_{x_1} - \mu_{x_2})})
\end{equation*}
where disk $1$ has radius $r_1$ and is centered at $x_1\thicksim\mathcal{N}(\mu_{x_1},\,\Sigma_{x_1})$; disk $2$ has radius $r_2$ and is centered at $x_2\thicksim\mathcal{N}(\mu_{x_2},\,\Sigma_{x_2})$; and $d = \|\mu_{x_1} - \mu_{x_2}\| - r_1 - r_2$ is the mean closest distance between the disks.
The ego dynamics are governed by the vehicle kinematic model and a model of the controller tracking error, assumed to be Gaussian distributed with covariance independent from the control sequence.
The world state represents the state of the ego vehicle and all other vehicles, and is assumed to be known at the current timestep, apart from other vehicles' intended motion patterns. Other vehicles' dynamics are modeled as RR-GP distributions~\cite{aoude_probabilistically_2013}, assumed to be independent from the ego vehicle state.  RR-GP provides mixture of Gaussian predictions that satisfy dynamic and environmental constraints. Each mixture component represents a motion pattern. For each motion pattern, RB-RHC uses PCL for the belief update $f_b^{\sim}$ in \eqref{eqn:p4_pcl}.

Problem (\ref{eqn:p4}) is solved by graph search in belief space. Given the RR-GP prediction models for all other vehicles, the belief of the world state can be fully determined by the spatial-temporal state of the ego vehicle $\{s_i, i\}$ independent of the control history. The state of the ego vehicle $s$ on the reference path can further be determined by the distance $\delta$ and velocity $\upsilon$ of the ego vehicle on the reference path. By discretizing the three dimensional space $\{\delta, \upsilon, i\}$, we obtain the node set $\mathcal{V}$ of the directed graph $\mathcal{G} = \{\mathcal{V}, \mathcal{E}\}$. An edge $\{\{\delta_1, \upsilon_1, i_1\}, \{\delta_2, \upsilon_2, i_2\}\}$ is in the edge set $\mathcal{E}$ if and only if $i_2 = i_1 + 1$ and there exists a legal acceleration profile (control $u$) to bring the vehicle from $\{\delta_1, \upsilon_1\}$ to $\{\delta_2, \upsilon_2\}$. Each edge binds with a cost based on the step-wise or final cost function and each node binds with a collision chance based on the world state belief represented by the node.

Without the chance constraint (\ref{eqn:p4_cc}), (\ref{eqn:p4}) can be solved by relaxing the edges in the graph $\mathcal{G}$ in the inverse temporal order. To address the chance constraint (\ref{eqn:p4_cc}), we introduce the Lagrange multiplier $\lambda$, and convert the constrained graph search to an unconstrained one by augmenting the cost of all edges with the collision chance bound with the corresponding end node multiplied by $\lambda$. Bisection search method is applied to find a value of $\lambda$ which corresponds to a locally optimal solution to \eqref{eqn:p4}~\cite{ono_chance-constrained_2015}.


\end{document}